\pdfoutput=1

\documentclass[11pt,a4paper,final]{article}
\usepackage[nohyperref]{acl2018}
\usepackage{times}
\usepackage{latexsym}
\usepackage{url}
\usepackage{booktabs}
\usepackage{subcaption}
\usepackage{listings}
\usepackage{color}
\usepackage{macros}
\usepackage{multirow}
\usepackage{booktabs, colortbl}
\usepackage{theorem}
\usepackage{enumitem}
\usepackage{stfloats}
\usepackage{amsmath}
\usepackage{amsfonts}
\usepackage{graphicx}
\usepackage{hyperref}
\usepackage{cleveref}

\newtheorem{definition}{Definition}

\newcommand{\ignore}[1]{}
\newcommand{\cameraready}[1]{#1}
\aclfinalcopy % Uncomment this line for the final submission
\def\aclpaperid{1159} %  Enter the acl Paper ID here

\newcommand\addsent{\textsc{AddSent}~}
\newcommand\wtq{\ms{WikiTableQuestions}~}

\title{Did the Model Understand the Question?}  

\author{Pramod K. Mudrakarta \\
  University of Chicago \\
  {\tt pramodkm@uchicago.edu} \\\And
  Ankur Taly \\
  Google Brain \\  
  \\\And
   Mukund Sundararajan \\
  Google \\ 
  {\tt \{ataly,mukunds,kedar\}@google.com} \\\And
   Kedar Dhamdhere \\
  Google \\  
  }

\date{}

\begin{document}
\maketitle
\begin{abstract}
We analyze state-of-the-art deep learning models for three tasks: question answering on (1) images, (2) tables, and (3) passages of text. Using the notion of \emph{attribution} \cameraready{(word importance)}, we find that these deep networks often ignore important question terms. Leveraging such behavior, we perturb questions to craft a variety of adversarial examples. Our strongest attacks drop the accuracy of a visual question answering model from $\vqaaccuracy$\% to $19\%$, and that of a tabular question answering model from $\npaccuracy$\% to $3.3\%$. Additionally, we show how attributions can strengthen attacks proposed by~\newcite{JL17} on paragraph comprehension models. Our results demonstrate that attributions can augment standard measures of accuracy and empower investigation of model performance. When a model is accurate but for the wrong reasons, attributions can surface erroneous logic in the model that indicates inadequacies in the test data.
\end{abstract}

\section{Introduction}

Recently, deep learning has been applied to a variety of question answering tasks. For instance, to answer questions about images (e.g.~\cite{kazemi2017show}), tabular data (e.g.~\cite{NLAMA17}), and passages of text (e.g.~\cite{YuDLLZC2018}). Developers, end-users, and reviewers (in academia) would all like to understand the capabilities of these models. 

The standard way of measuring the goodness of a system is to evaluate its error on a test set. High accuracy is indicative of a good model only if the test set is representative of the underlying real-world task. Most tasks have large test and training sets, and it is hard to manually check that they are representative of the real world. 

In this paper, we propose techniques to analyze the sensitivity of a deep learning model to question words. We do this by applying \emph{attribution} (as discussed in~\cref{sec:prelim}), and generating adversarial questions. Here is an illustrative example: recall Visual Question Answering~\cite{agrawal2015vqa} where the task is to answer questions about images. Consider the question ``\textit{how symmetrical are the
white bricks on either side of the building?}'' (corresponding image in Figure~\ref{fig:vqa:example}). The system that we study gets the answer right (``\textit{very}''). But, we find (using an attribution approach) that the system relies on only a few of the words like ``how'' and ``bricks''. Indeed, we can construct adversarial questions about the same image that the system gets wrong. For instance, \cameraready{``\textit{how spherical are the white bricks on either side of the building?}''} returns the same answer (``\textit{very}''). A key premise of our work is that most humans have expertise in question answering. Even if they cannot manually check that a dataset is representative of the real world, they can identify important question words, and anticipate their function in question answering.

\subsection{Our Contributions}
We follow an analysis workflow to understand three question answering models. There are two steps. First, we apply Integrated Gradients~(henceforth, IG)~\cite{STY17} to attribute the 
systems' predictions to words in the questions. We propose visualizations of attributions to make
analysis easy. Second, we identify weaknesses (e.g., relying on unimportant words) in the networks' logic as exposed by the attributions,
and leverage them to craft adversarial questions.

A key contribution of this work is an
  \emph{overstability} test for question answering networks.
  \newcite{JL17} showed that reading comprehension networks are overly
  stable to semantics-altering edits to the passage. In this work, we
  find that such overstability also
  applies to questions. Furthermore, this behavior can be seen in visual
  and tabular question answering networks as well. We use attributions
  to a define a general-purpose test for measuring the extent of the
  overstability (\cref{sec:np:overstability,sec:vqa:overstability}). It involves measuring how a network's accuracy
  changes as words are systematically dropped from questions.

\cameraready{We emphasize that, in contrast to model-independent
  adversarial techniques such as that of ~\newcite{JL17}, our method
  exploits the strengths and weaknesses of the model(s) at hand. This
  allows our attacks to have a high success rate. Additionally, using insights
  derived from attributions we were able to improve the attack success rate of~\newcite{JL17} (\cref{sec:squad:attacks}). Such
  extensive use of attributions
  in crafting adversarial examples is novel to the best of our
  knowledge.}

Next, we provide an overview of our results. \cameraready{In each case, we evaluate a pre-trained model
on new inputs. We keep the networks' parameters intact.}

\begin{description}[leftmargin=*]
%We found that NP often ignores question-terms. A large portion of the signal comes 
%directly from the table. For instance, we found that in tables that were about medal counts in the Olympics, NP was predisposed to choose a program that picks a row that is the ``\opmax'' over a column. Second, NP often gets the question right for the wrong reasons. For instance, for the question
%``which nation earned the most gold medals'', the operations selected by NP are ``\opfirst''\ (pick the first row of the table).  It gets the answer right, but only because the table
%happens to be arranged in order of rank. We quantify this weakness by reordering the rows of the tables; this causes NP's accuracy on the dev dataset to drop from $\npaccuracy$  to $29.05\%$. We find that NP often relies on content-free words. We drop such words (e.g., ``a'', ``at'', ``the'', etc.) from the question, and accuracy falls from $\npaccuracy$ to
%$28.61\%$. Finally, NP over-relies on certain trigger words without comprehending grammar; when we prefix or suffix the 
%question with a semantically irrelevant phrase like ``in not a lot of words''. We find that this 
%drops the accuracy of NP. The union of $12$ such attacks drops its accuracy on the dev dataset from $\npaccuracy$ to $5.01\%$. 
%\todo{ankur: redo the NP section.}

\item[\textbf{Visual QA (\cref{sec:vqa}):}] The task is to
  answer questions about images.  We analyze the deep network
  in~\newcite{kazemi2017show}.  We find that the network
    ignores many question words, relying largely on the image to
    produce answers. For instance, we show that the model retains more
    than 50\% of its original accuracy even when every word that is not ``color'' is deleted from all questions in the validation set. We also show that the model under-relies on important
    question words (e.g. nouns) and attaching content-free prefixes (e.g., ``in not
    many words, $\ldots$") to questions drops the accuracy from
     $\vqaaccuracy$\% to 19\%.

\item[\textbf{QA on tables (\cref{sec:tableqa}):}] We analyze
  a system called Neural Programmer (henceforth, NP)~\cite{NLAMA17} that answers questions on tabular data.  \cameraready{NP determines the answer
    to a question by selecting a sequence of operations to apply on
    the accompanying table (akin to an SQL query; details in~\cref{sec:tableqa})}. We
    find that these operation selections are more influenced by
    content-free words (e.g., ``in'', ``at'', ``the'', etc.) in
    questions than important words such as nouns or
    adjectives. Dropping all content-free words reduces the validation
    accuracy of the network from $\npaccuracy\%$\footnote{This is the
    single-model accuracy that we obtained on training the Neural
    Programmer network. The accuracy reported in the paper is 34.1\%.}
  to $28.5\%$. Similar to Visual QA, we show that
  attaching content-free phrases (e.g., ``in not a lot of words'') to the
  question drops the network's accuracy from $\npaccuracy\%$
  to $3.3\%$. We
  also find that NP often gets the answer right for the wrong
  reasons. For instance, for the question ``\textit{which nation
    earned the most gold medals?}'', one of the operations selected by
  NP is ``\opfirst''\ (pick the first row of the table).  Its
  answer is right only because the table happens to be arranged in order
  of rank.  We quantify this weakness by evaluating NP on the set of
  perturbed tables generated by~\newcite{pasupat2016inferring} and
  find that its accuracy drops from
  $\npaccuracy\%$ to $23\%$. Finally, we show an extreme form of overstability where the table itself
  induces a large bias in the network regardless of the question. For
  instance, we found that in tables about Olympic medal counts, NP was predisposed to selecting the
  \cameraready{``$\opprev$''} operator.

\item[\textbf{Reading comprehension (Section~\ref{sec:rc}):}] The task
  is to answer questions about paragraphs of text.  We analyze the
  network by~\newcite{YuDLLZC2018}. Again, we find that the network
  often ignores words that should be important.
  \newcite{JL17} proposed attacks wherein sentences are added to
  paragraphs that ought not to change the \cameraready{network's}
  answers, but sometimes do. Our main finding is that
  these attacks are more likely to succeed when an added sentence
  includes all the question words that the model found important
  (for the original paragraph). For instance, we find that attacks are 50\% more likely to be successful when the added sentence includes top-attributed nouns in the
  question. This insight should
  allow the construction of more successful attacks and better
  training data sets.
\end{description}

In summary, we find that all networks ignore important parts of questions.  One can fix this by either improving training
data, or introducing an inductive bias.  Our analysis workflow is
helpful in both cases. It would also make sense to expose end-users to
attribution visualizations. Knowing which words were ignored, or which
operations the words were mapped to, can help the user decide whether
to trust a system's response.

\section{Related Work}
We are motivated by~\newcite{JL17}. As they discuss, ``\textit{the
  extent to which [reading comprehension systems] truly understand
  language remains unclear}''.  The contrast between~\newcite{JL17}
and our work is instructive.  Their main contribution is to fix the
evaluation of reading comprehension systems by augmenting the test set
with adversarially constructed examples. (As they point out in
Section 4.6 of their paper, this does not necessarily fix the model;
the model may simply learn to circumvent the specific attack underlying the adversarial
examples.)  \cameraready{Their method is independent of the specification of the model at hand. They use crowdsourcing to craft passage perturbations intended
  to fool the network, and then query the network to test their
  effectiveness.}

In contrast, we propose improving the analysis of
question answering systems. Our method peeks into the
  logic of a network to identify high-attribution question
  terms. Often there are several important question terms (e.g.,
  nouns, adjectives) that receive tiny attribution. We leverage this
  weakness and perturb questions to craft targeted attacks.
While~\newcite{JL17} focus exclusively on systems for the reading
comprehension task, we analyze one system each for three different
tasks. Our method also helps improve the efficacy~\newcite{JL17}'s attacks; see
  \cref{appendix:tbl:modified_squad_attacks} for examples.
%Also, their attacks are constructed on the passage, ours are on the question.
Our analysis technique is specific to deep-learning-based systems,
whereas theirs is not.

We could use many other methods instead of Integrated Gradients (IG)
to attribute a deep network's prediction to its input
features~\cite{BSHKHM10,SVZ13,SGSK16,BMBMS16,SDBR14}. One could also use
model agnostic techniques like~\newcite{ribeiro2016should}. We choose IG for
its ease and efficiency of implementation (requires just a
few gradient-calls) and its axiomatic justification (see
~\newcite{STY17} for a detailed comparison with other attribution
methods).

Recently, there have been a number of techniques for crafting and
defending against adversarial attacks on image-based deep learning
models (cf.~\newcite{adversarial}). They are based on
oversensitivity of models, i.e., tiny, imperceptible
perturbations of the image to change a model's response. In
contrast, our attacks are based on models' over-reliance on few
question words even when other words should matter.

We discuss task-specific related work in corresponding sections
(\cref{sec:vqa,sec:tableqa,sec:rc}).

\section{Integrated Gradients (IG)}
\label{sec:prelim}
We employ an attribution technique called Integrated
Gradients (IG)~\cite{STY17} to isolate question words that a deep
learning system uses to produce an answer.

Formally, suppose a function $F: \reals^n \rightarrow [0,1]$ represents a deep network, and an input $x = (x_1,\ldots,x_n) \in
\reals^n$.  An attribution of the prediction at input $x$ relative to
a baseline input $\xbase$ is a vector $A_F(x, \xbase) =
(a_1,\ldots,a_n) \in \reals^n$ where $a_i$ is the \emph{contribution}
of $x_i$ to the prediction $F(x)$. One can think of $F$ as the
probability of a specific response.
$x_1,\ldots,x_n$ are the question words; to be precise, they are
going to be vector representations of these terms. The attributions
$a_1,\ldots,a_n$ are the influences/blame-assignments to the
variables $x_1,\ldots,x_n$ on the probability $F$.

Notice that attributions are defined relative to a special, uninformative
input called the \emph{baseline}. In this paper, we use an empty question as the
baseline, that is, a sequence of word embeddings corresponding to padding value. Note that the context (image, table, or passage) of the baseline $\xbase$ is set to be that of $x$; only the question is set to empty. We now describe how IG produces attributions.

Intuitively, as we interpolate between the baseline and the input, the prediction moves 
along a trajectory, from uncertainty to certainty (the final probability).
At each point on this trajectory, one can use the gradient of the function $F$ with
respect to the input to attribute the change in probability back to the input variables.
IG simply aggregates the gradients of the probability with respect to the
input along this trajectory using a path integral.
%% \cameraready{Specifically, integrated gradients
%% are defined as the path integral of the gradients along the straightline
%% path from the baseline $\xbase$ to the input $x$.}

\begin{definition}[\textbf{Integrated Gradients}]\label{def:intgrad}
Given an input $x$ and baseline
$\xbase$, the integrated gradient along the $i^{th}$ dimension is defined as follows.
%\begin{equation}\label{eqn:intgrad}
\[\integratedgrads_i(x,\xbase) \synteq (x_i-\xbase_i)\times\int_{\sparam=0}^{1} \tfrac{\partial F(\xbase + \sparam\times(x-\xbase))}{\partial x_i  }~d\sparam\]
%\end{equation}
(here
$\tfrac{\partial F(x)}{\partial x_i}$ is the gradient of
$F$ along the $i^{th}$ dimension at $x$).
\end{definition}

\newcite{STY17} discuss several properties of IG. Here, we informally mention a few desirable ones, deferring the reader to~\newcite{STY17} for formal definitions.
 
\cameraready{IG satisfies the condition that the attributions sum to
  the difference between the probabilities at the input and the
  baseline. We call a variable uninfluential if all else fixed,
  varying it does not change the output probability. IG satisfies the
  property that uninfluential variables do not get any
  attribution. Conversely, influential variables always get some
  attribution. Attributions for a linear combination of two
  functions $F_1$ and $F_2$ are a linear
  combination of the attributions for $F_1$ and $F_2$. Finally, IG
  satisfies the condition that symmetric variables get equal
  attributions.}

\cameraready{In this work, we validate the use of IG empirically via question
  perturbations. We observe that perturbing high-attribution terms changes the networks' response (\cref{sec:vqa:attacks,sec:np:attacks}). Conversely, perturbing terms that receive a
  low attribution does not change the network's response (\cref{sec:vqa:overstability,sec:np:overstability}). We use these observations to craft
  attacks against the network by perturbing instances where generic words (e.g., ``a'', ``the'') receive high attribution or contentful words receive low attribution.}

\section{Visual Question Answering}
\label{sec:vqa}

\subsection{Task, model, and data}	
The Visual Question Answering Task~\cite{agrawal2015vqa,teney2017tips,kazemi2017show,ben2017mutan,zhu2016visual7w} requires a system to answer questions about images (\cref{fig:vqa:example}). We analyze the deep network from~\newcite{kazemi2017show}. It achieves $\vqaaccuracy$\% accuracy on the validation set (the state of the art~\cite{fukui2016multimodal} achieves $66.7\%$). We chose this model for its easy reproducibility. 

The VQA 1.0 dataset~\cite{agrawal2015vqa} consists of 614,163 questions posed over 204,721 images (3 questions per image). The images were taken from COCO~\cite{lin2014microsoft}, and the questions and answers were crowdsourced.

The network in \newcite{kazemi2017show} treats question answering as a classification task wherein the classes are 3000 most frequent answers in the training data. The input question is tokenized, embedded and fed to a multi-layer LSTM. The states of the LSTM attend to a featurized version of the image, and ultimately produce a probability distribution over the answer classes.

\subsection{Observations}
We applied IG and attributed the top selected answer class 
to input question words. The baseline for a given input instance is the image and an empty question\footnote{We do not black out the image in our baseline as our objective is to study the influence of just the
question words for a given image}. 
We omit instances where the top answer class predicted by the network remains the same even when the
question is emptied (i.e., the baseline input). 
This is because IG attributions are not informative when the input
and the baseline have the same prediction.
%We do not compute attributions when an empty question (the baseline for IG) passed to the image results
%in the same answer as when the question is passed.

\begin{figure}
    \begin{minipage}{0.25\linewidth}
        \includegraphics[width=\linewidth]{COCO_val2014_000000058655.jpg} 
    \end{minipage}\hspace{5pt}\begin{minipage}{0.75\linewidth}
        Question: {\color[rgb]{0.996094,0.250000,0.250000}how} {\color[rgb]{0.500000,0.500000,0.500000} symmetrical} {\color[rgb]{0.683594,0.406250,0.406250} are} {\color[rgb]{0.542969,0.476562,0.476562} the} {\color[rgb]{0.441406,0.441406,0.617188} white} {\color[rgb]{0.617188,0.441406,0.441406} bricks} {\color[rgb]{0.472656,0.472656,0.554688} on} {\color[rgb]{0.378906,0.378906,0.738281} either} {\color[rgb]{0.480469,0.480469,0.539062} side} {\color[rgb]{0.484375,0.484375,0.531250} of} {\color[rgb]{0.515625,0.492188,0.492188} the} {\color[rgb]{0.511719,0.492188,0.492188} building} \\
        Prediction: very \\
        Ground truth: very
    \end{minipage} \\
    \caption{\small Visual QA~\cite{kazemi2017show}: Visualization of attributions (word importances) for a question that the network gets right. Red indicates high attribution, blue negative attribution, and gray near-zero attribution. The colors are determined by attributions normalized w.r.t the maximum magnitude of attributions among the question's  words.}\label{fig:vqa:example}
\end{figure}

A visualization of the attributions is shown in \cref{fig:vqa:example}. 
Notice that very few words have high attribution.
We verified that altering the low attribution words in the question does not change the network's
 answer. For instance, the following questions still return ``\textit{very}'' as the answer:  ``\textit{how spherical are the white bricks on either side of the building}'', 
``\textit{how soon are the bricks fading on either side of the building}'', 
``\textit{how fast are the bricks speaking on either side of the building}''. 

On analyzing attributions across examples, we find that most of the highly attributed words are words such as ``\textit{there}'',  ``\textit{what}'', ``\textit{how}'', ``\textit{doing}''-- they are usually the less important words in questions. In \cref{sec:vqa:overstability} we describe
a test to measure the extent to which the network depends on such words.
We also find that informative words in the question (e.g., nouns) often receive very low
attribution, indicating a weakness on part of the network. In Section~\ref{sec:vqa:attacks},
we describe various attacks that exploit this weakness.

\subsection{Overstability test}
\label{sec:vqa:overstability}
To determine the set of question words that the network finds most important, we isolate words
that most frequently occur  as top attributed words in questions. We then drop
all words except these and compute the accuracy.

\begin{figure}[h]
    \includegraphics[width=\linewidth]{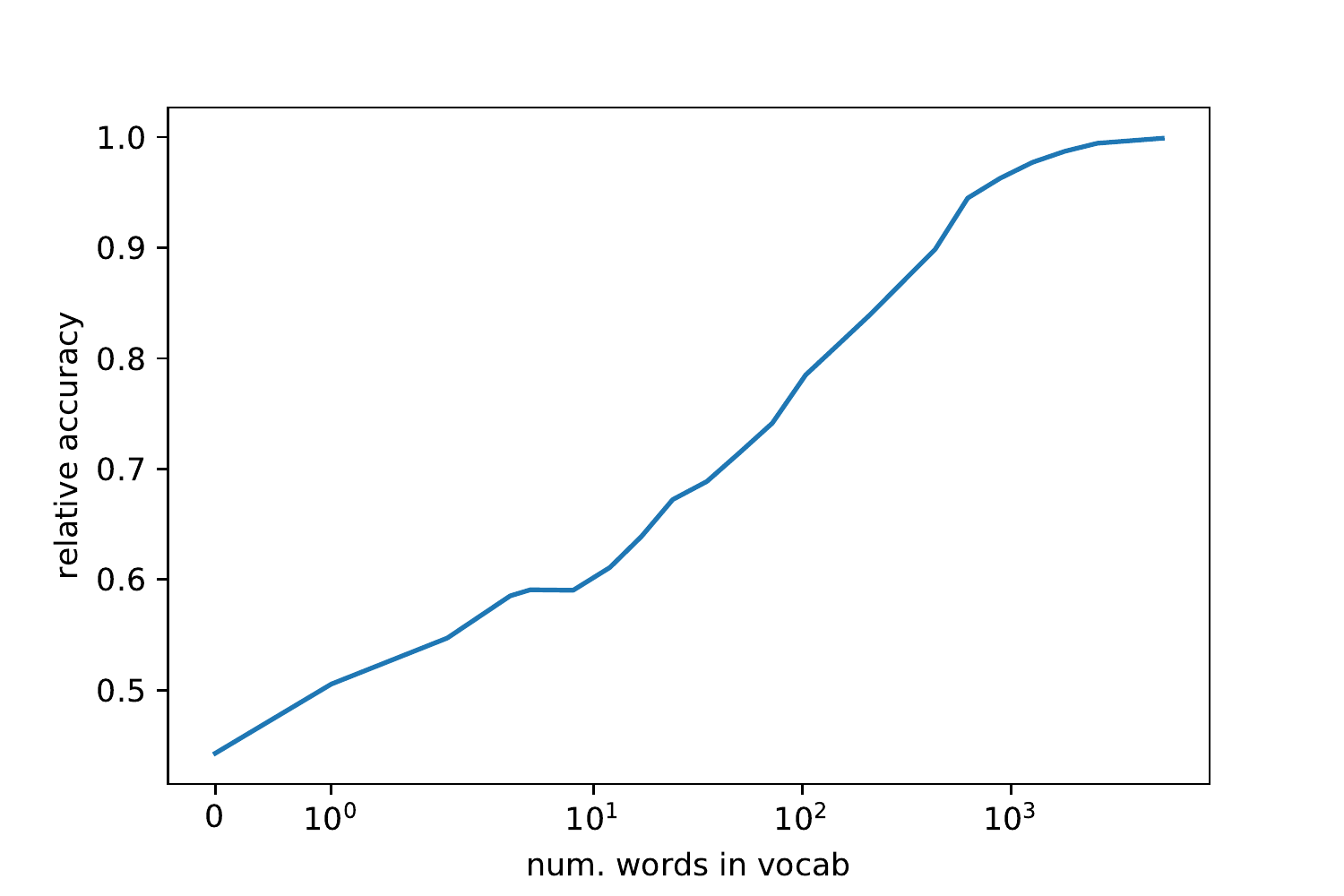}
    \caption{\small VQA network~\cite{kazemi2017show}: Accuracy as a function of vocabulary size, relative to its original accuracy. Words are chosen in the descending order of how frequently they appear as top attributions. The X-axis is on logscale, except near zero where it is linear.} \label{fig:vqa:whitelist_attack_curve}
\end{figure}

\Cref{fig:vqa:whitelist_attack_curve} shows how the accuracy changes
as the size of this isolated set is varied from 0 to 5305. We find that just one word is enough for the model to achieve more than 50\% of its final accuracy. That word is ``color''. 

\cameraready{Note that even when empty questions are passed as input
  to the network, its accuracy remains at about 44.3\% of its original
  accuracy. This shows that the model is largely reliant on the image
  for producing the answer.}
    
\cameraready{The accuracy increases (almost) monotonically with the
  size of the isolated set. The top 6 words in the isolated set are
  ``color'', ``many'', ``what'', ``is'', ``there'', and ``how''.  We
  suspect that generic words like these are used to determine the type
  of the answer. The network then uses the type to choose between a
  few answers it can give for the image.}

%One can construct attacks by introducing words 
%in the fragmented questions which have semantic relevance to the image, 
%but are rarely used by the network in predicting the answer. Doing this at scale 
%would require crowdsourcing to check for grammatical correctness of the questions.
%See \ref{sec:vqa:subjectattacks} for a similar kind of attack.

%This shows that the model relies on generic words, i.e., words appearing in commonly in several questions, to determine the answer. From inspection, we suspect that these generic words are used to determine the type of the answer and the network simply picks from one of the few answers it can give, based on this type. Further, we can construct attacks by introducing words in the fragmented questions which have semantic relevance to the image, but are rarely used by the network in predicting the answer. To do this at scale would require crowdsourcing to check for grammatical correctness of the questions. See \ref{sec:vqa:subjectattacks} for a similar kind of attack.

\subsection{Attacks}\label{sec:vqa:attacks}
Attributions reveal that the network relies largely on generic words in answering questions (\cref{sec:vqa:overstability}).
This is a weakness in the network's logic. We now describe a few attacks against the network that exploit this weakness. 

\subsubsection*{Subject ablation attack}
In this attack, we replace the subject of a question with a specific noun that consistently receives low attribution across questions. \cameraready{We then determine, among the questions that the network originally answered correctly, what percentage result in the same answer after the ablation. We repeat this process for different nouns; specifically, ``fits'', ``childhood'', ``copyrights'', ``mornings'', ``disorder'', ``importance'', ``topless'', ``critter'', ``jumper'', ``tweet'', and average the result.}

\cameraready{We find that, among the set of questions that the network originally answered correctly, 75.6\%
of the questions return the same answer despite the subject replacement.}

\subsubsection*{Prefix attack}
In this attack, we attach content-free phrases to questions. The phrases are manually crafted using 
generic words that the network finds important (\cref{sec:vqa:overstability}). \Cref{tbl:vqa:prefix_attacks} (top half) shows the 
resulting accuracy for three prefixes ---``\textit{in not a lot of words}'', ``\textit{what is the answer to}'',
and ``\textit{in not many words}''.  All of these phrases nearly halve the model's accuracy. The union of the three attacks drops the model's accuracy from \cameraready{ $\vqaaccuracy$\% to 19\%}.
%All of these 
%such that they are content-free but contain 
% to
%construct prefix phrases which do not alter
%We construct prefix phrases which do not alter the meaning of the question, but are designed to modify the network’s answer. We computed the accuracy on a sample of 5000 questions from the validation set. The results are shown in Table~\ref{tbl:vqa:prefix_attacks}. To confuse the network to think of question types (1), (2) as type (3), we prefix “what is the answer to” to the question. To confuse the network to think of question types (1), (3) as (2), we append “in not many words” to the question. All of these attacks nearly halves the model's accuracy.

We note that the attributions computed for the network were crucial in crafting the prefixes.
For instance, we find that other prefixes like ``\textit{tell me}'', ``\textit{answer this}'' and ``\textit{answer this for me}''
do not drop the accuracy by much; see \cref{tbl:vqa:prefix_attacks} (bottom half). The union of these three ineffective prefixes drops the accuracy from $\vqaaccuracy$\% to only 46.9\%.
Per attributions, words present in these prefixes are not deemed important by the network.
%Furthermore, to ensure that the drop in accuracy does not result from merely the question length, 
%we append the first five words of each question to itself and note that the accuracy does not drop by
%more than 7\%. Based on attributions, the prefix “tell me”, “answer this”, should not alter the answer, and we verified that the accuracy does not drop by more than 7\% with these prefixes. 

\begin{table}
    \centering \small
\begin{tabular}{lc}
    \toprule
    Prefix & Accuracy \\
    \midrule
in not a lot of words &    35.5\% \\
in not many words &    32.5\% \\
what is the answer to &    31.7\% \\
\midrule
    \emph{Union of all three}	& \textbf{19\%} \\
    \midrule
    Baseline prefix & \\
    \midrule
               tell me &    51.3\% \\
answer this &    55.7\% \\
answer this for me &    49.8\% \\
\midrule
\emph{Union of baseline prefixes} & \textbf{46.9\%} \\
    \bottomrule
\end{tabular}

\caption{\small VQA network~\cite{kazemi2017show}: Accuracy for prefix attacks; original accuracy is $\vqaaccuracy$\%.}
    \label{tbl:vqa:prefix_attacks}
\end{table}

\subsection{Related work}
\newcite{agrawal2016analyzing} analyze several VQA models. Among other attacks,
they test the models on question fragments of telescopically increasing length. They observe
that VQA models often arrive at the same answer by looking at a small fragment of
the question. Our stability analysis in \cref{sec:vqa:overstability}
explains, and intuitively subsumes this; indeed, several of the top attributed words
appear in the prefix, while important words like ``color'' often occur in the
middle of the question. Our analysis enables additional attacks, for instance,
replacing question subject with low attribution nouns.
\newcite{ribeiro2016nothing} use a model explanation
technique to illustrate overstability for two examples. They do not quantify
their analysis at scale.  \newcite{kafle2017analysis, zhang2016yin}
examine the VQA data, identify deficiencies, and propose data augmentation
to reduce over-representation of certain question/answer types.
\newcite{goyal2016making} propose the VQA 2.0 dataset, which has pairs of similar
images that have different answers on the same question. We note that our
method can be used to improve these datasets by identifying inputs where models
ignore several words. \newcite{huang2017novel} evaluate robustness of VQA models
by appending \emph{questions} with semantically similar questions. Our prefix
attacks in \cref{sec:vqa:attacks} are in a similar vein and perhaps a more
natural and targeted approach.
Finally, \newcite{fong2017interpretable} use saliency methods to produce image
perturbations as adversarial examples; our attacks are on the question.

\section{Question Answering over Tables}
\label{sec:tableqa}

\subsection{Task, model, and data}
\label{sec:tableqa_intro}
We now analyze question answering over tables based on the \wtq benchmark
dataset~\cite{Pasupat15compositionalsemantic}. The dataset has $22033$
questions posed over $2108$ tables scraped from Wikipedia. Answers are either contents of table cells or some table aggregations. Models performing QA on tables translate the question into a structured program (akin to an SQL query) which is then executed on the table to produce the answer. We analyze a model
called Neural Programmer (NP)~\cite{NLAMA17}. NP is the state of the art among
models that are weakly supervised, i.e., supervised using the final answer
instead of the correct structured program. It achieves $\npaccuracy$\%
accuracy on the validation set.

NP translates the input into a structured program consisting of four operator and table column selections. An example of such a program is ``\opresetselect\ (score), \opresetselect\ (score), \opmin\ (score), \opprint\ (name)'', where the output is the name of the person who has the lowest score. 

\subsection{Observations}
We applied IG to attribute operator and column selection to question words. NP preprocesses inputs and whenever applicable, appends symbols $\entrymatchtok, \colmatchtok$ to questions that signify matches between a question and the accompanying table. These symbols are treated the same as question words. NP also computes priors for column selection using question-table matches. These vectors, $\tblmatchbias$ and $\colmatchbias$, are passed as additional inputs to the neural network. In the baseline for IG, we use an empty question, and zero vectors for column selection priors\footnote{Note that the table is left intact in the baseline}. 

\begin{figure}[h]
    \centering
    \includegraphics[width=\linewidth]{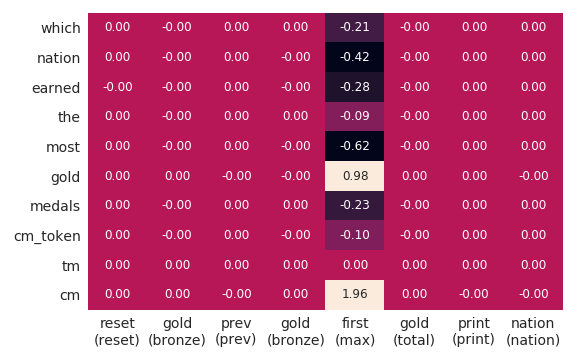}
    \caption{\small \cameraready{Visualization of attributions. Question words, preprocessing tokens and column selection priors on the Y-axis. Along the X-axis are operator and column selections with their baseline counterparts in parentheses. Operators and columns not affecting the final answer, and those which are same as their baseline counterparts, are given zero attribution. }}
    \label{fig:np:visualization}
\end{figure}

We visualize the attributions using an alignment matrix; they are commonly used in
the analysis of translation models (\cref{fig:np:visualization}).
Observe that the operator ``\opfirst'' is used when the question is asking for a
superlative. Further, we see that the word
``gold'' is a trigger for this operator. We investigate implications of this behavior in the following sections.

\subsection{Overstability test}\label{sec:np:overstability}
\begin{figure}
    \includegraphics[width=\linewidth]{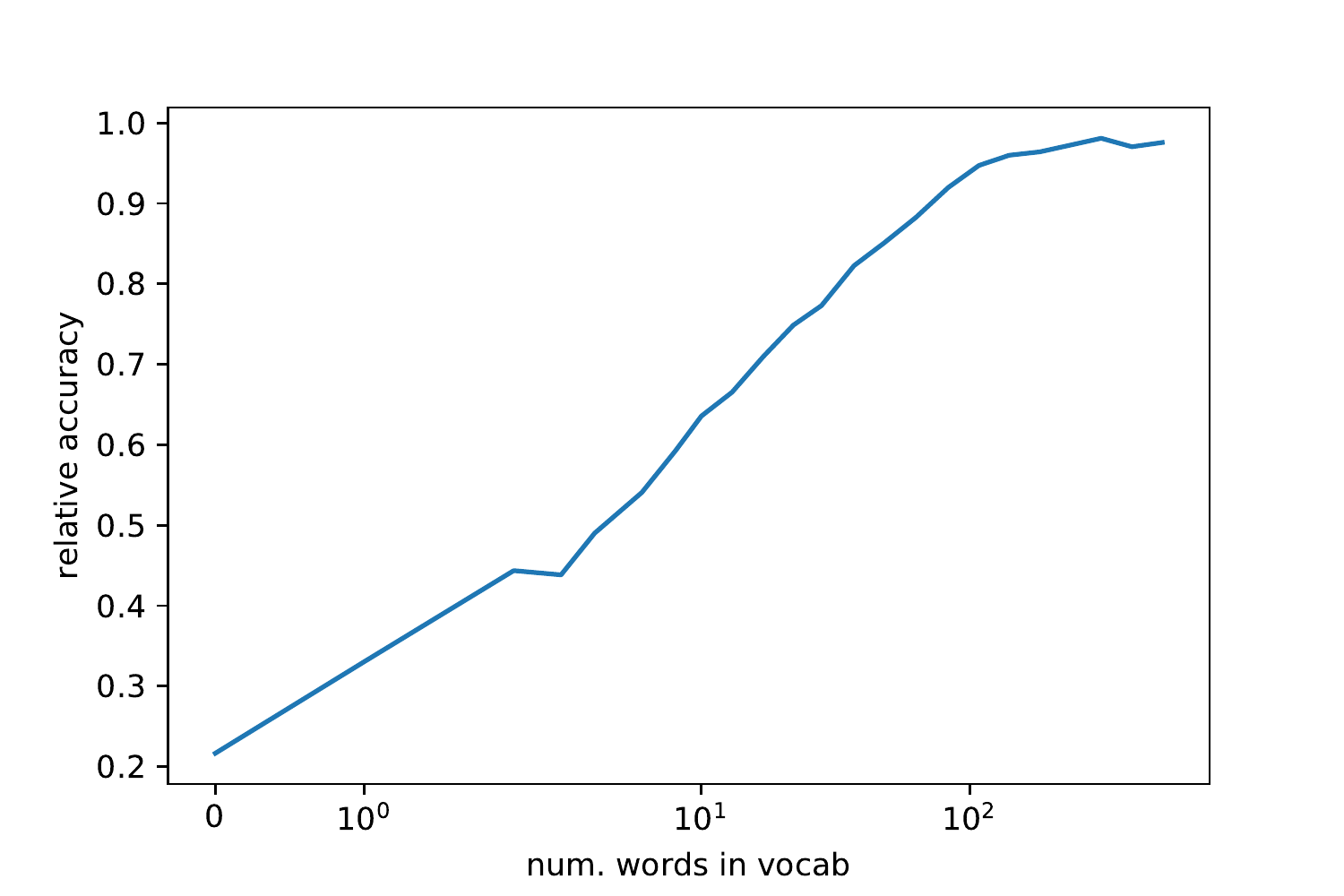}
    \caption{\small Accuracy as a function of vocabulary size. The words are chosen in the descending order of their frequency appearance as top attributions to question terms. The X-axis is on logscale, except near zero where it is linear. Note that just 5 words are necessary for the network to reach more than 50\% of its final accuracy.}
    \label{fig:np:overstability}
\end{figure}

Similar to the test we did for Visual QA (\cref{sec:vqa:overstability}), we check for overstability in NP by
looking at accuracy as a function of the vocabulary size. We treat table match annotations \tblmatch, \colmatch\ and the out-of-vocab token ($\mathit{unk}$) as part of the vocabulary. The results are in \cref{fig:np:overstability}. We see that the curve is similar to that of Visual QA (\cref{fig:vqa:whitelist_attack_curve}). Just 5 words (along with the column selection priors) are sufficient for
the model to reach more than 50\% of its final accuracy on the validation
set. These five words are: ``many'', ``number'', ``$\tblmatch$'', ``after'', and ``total''.

\subsection{Table-specific default programs}
We saw in the previous section that the model relies on only a few words in producing correct answers. An extreme case of overstability is when the operator sequences produced by the model are independent of the question. We find that if we supply an empty question as an input, i.e., the output is a function only of the table, then the distribution over programs is quite skewed. We call these programs table-specific default programs. On average, about $36.9\%$ of the selected operators match their table-default counterparts, indicating that the model relies significantly on the table for producing an answer.

For each default program, we used IG to attribute operator and column selections to column names and show ten most frequently occurring ones across tables in the validation set (\cref{tbl:default_programs}). 

Here is an insight from this analysis: NP uses the combination ``\opresetselect, \opprev'' to exclude the last row of the table from answer computation. The default program corresponding to ``\opresetselect, \opprev, \opmax, \opprint'' has attributions to column names such as \textit{``rank'', ``gold'',
    ``silver'', ``bronze'', ``nation'', ``year''}. These column names indicate medal tallies and usually have a \textit{``total''} row.
If the table happens not to have a \textit{``total''} row, the model may produce an incorrect answer.

\begin{table*}[!htbp]
    \centering \small 
    \begin{tabular}{lrll}
        \toprule
        Operator sequence &  \# &                                                                              Triggers & Insights\\
        \midrule
        
        \opresetselect, \opresetselect, \opmax, \opprint &      109 &                     [\unktoken, date, position, points, name, competition, notes, no, year, venue] & sports\\
        \opresetselect, \opprev, \opmax, \opprint &       68 &                             [\unktoken, rank, total, bronze, gold, silver, nation, name, date, no] & medal tallies\\
        \opresetselect, \opresetselect, \opfirst, \opprint &       29 &              [name, \unktoken, notes, year, nationality, rank, location, date, comments, hometown] & player rankings\\
        \opresetselect, \opgroupbymax, \opfirst, \opprint &       25 &                         [notes, date, title, \unktoken, role, genre, year, score, opponent, event] & awards\\
        \opresetselect, \opresetselect, \opmin, \opprint &       17 &                            [year, height, \unktoken, name, position, floors, notes, jan, jun, may] & building info.\\
        \opresetselect, \opgroupbymax, \opmax, \opprint &       14 &             [opponent, date, result, location, rank, site, attendance, notes, city, listing] & politics\\
        \opresetselect, \opnext, \opfirst, \opprint &       10 &                             [\unktoken, name, year, edition, birth, death, men, time, women, type] & census\\
        \bottomrule
    \end{tabular}
    \caption{\small Attributions to column names for table-specific default programs  (programs returned by NP on empty input questions). See supplementary material, \cref{appendix:tbl:default_programs} for the full list. These results are indication that the network is predisposed towards picking certain operators solely based on the table.}
    \label{tbl:default_programs}
    \vspace{-0.5cm}
\end{table*}

We now describe attacks that add or drop content-free words
from the question, and cause NP to produce the wrong answer.
These attacks leverage the attribution analysis. 

\subsection{Attacks}\label{sec:np:attacks}
\subsubsection*{Question concatenation attacks}
In these attacks, we either suffix or prefix content-free phrases to questions. The phrases
are crafted using \cameraready{irrelevant} trigger words for operator selections (supplementary material, \cref{appendix:tbl:operatortriggers}). We manually ensure that the phrases are content-free. 

\begin{table}
    \centering `
    \small
    \begin{tabular}{lll}
        \toprule
        Attack phrase & Prefix & Suffix \\
        \midrule
        in not a lot of words &  20.6\% &  10.0\% \\
        if its all the same &  21.8\% &  18.7\% \\
        in not many words &  15.6\% &  11.2\% \\
        one way or another &  23.5\% &  20.0\% \\
        \midrule
        \textit{Union of above attacks} & \multicolumn{2}{c}{\textbf{3.3\%}} \\
        \midrule
        Baseline & \\
        \midrule
        please answer &  32.3\% &  30.7\% \\
        do you know &  31.2\% &  29.5\% \\
        \midrule
        \textit{Union of baseline prefixes} & \multicolumn{2}{c}{\textbf{27.1\%}} \\
        \bottomrule
    \end{tabular}
    \caption{\small Neural Programmer~\cite{NLAMA17}: Left: Validation accuracy when attack phrases are concatenated to the question. (Original: $\npaccuracy$\%)}
    \label{tbl:prefixattacks}
    \label{tbl:rowperturb}
\end{table}

\Cref{tbl:prefixattacks} describes our results. The first $4$ phrases use irrelevant trigger words and result in a large drop in accuracy. For instance, the first phrase uses ``not'' which is a trigger for ``\opnext'', ``\oplast'', and ``\opmin'', and the second uses ``same'' which is a trigger for ``\opnext'' and ``\opgroupbymax''. The four phrases combined results in the model's accuracy going down from $\npaccuracy$\% to $3.3\%$. The first two phrases alone drop the accuracy to $5.6\%$.

The next set of phrases use words that receive low attribution across questions, and are hence non-triggers for any operator. The resulting drop in accuracy on using these phrases is relatively low. Combined, they result in the model's accuracy dropping from $\npaccuracy\%$ to $27.1\%$.

\subsubsection*{Stop word deletion attacks}
We find that sometimes an operator is selected based on \emph{stop words} like: ``a'', ``at'', ``the'', etc.
For instance,  in the question, ``what ethnicity is at the top?'', the operator
``$\opnext$'' is triggered on the word ``at''. Dropping the word ``at'' from
the question changes the operator selection and causes NP to
return the wrong answer.

We drop stop words from questions in the validation dataset that were
originally answered correctly and test NP on them. The stop words to
be dropped were manually selected\footnote{We avoided standard
  stop word lists (e.g. NLTK) as they contain contentful
  words (e.g ``after'') which may be important in some questions (e.g. ``who ranked right after turkey?'')} and
are shown in Figure~\ref{fig:fluffwords} in the supplementary material.

By dropping stop words, the accuracy drops from \cameraready{$\npaccuracy$\% to $28.5\%$}. Selecting operators based on stop words is not robust.
In real world search queries, users often phrase questions without stop words, trading  grammatical correctness for conciseness. For instance, the user may simply say ``top ethnicity''. It may be possible to defend against such examples by generating synthetic training data, and re-training the network on it.

\subsubsection*{Row reordering attacks} 
\label{sec:np:reorderingattacks}
We found that NP often got the question right by leveraging artifacts of the table.
For instance, the operators for the question ``which nation earned the most gold medals'' are ``\opresetselect'', ``\opprev'', ``\opfirst'' and ``\opprint''. The ``\opprev'' operator essentially excludes the last row from the answer computation. It gets the answer right for two reasons: (1) the answer is not in the last row, and (2) rows are sorted by the values in the column ``gold''. 

In general, a question answering system should not rely on row
ordering in tables. To quantify the extent of such biases,
we used a perturbed version of \wtq validation dataset as described
in \newcite{pasupat2016inferring}\footnote{based on data at
  \url{https://nlp.stanford.edu/software/sempre/wikitable/dpd/}} and
evaluated the existing NP model on it (there was no re-training
involved here). We found that NP has only $23\%$ accuracy on it, in constrast to
an accuracy of $\npaccuracy$\% on the original validation dataset.

One approach to making the network robust to row-reordering
  attacks is to train against perturbed tables. This may also help the
  model generalize better. Indeed, \newcite{mudrakarta2018training} note that
  the state-of-the-art strongly supervised\footnote{supervised on the structured program} model on \wtq~\cite{krishnamurthy2017neural} enjoys
  a  $7\%$ gain in its final accuracy by leveraging perturbed tables during training.

\section{Reading Comprehension}
\label{sec:rc}
\definecolor{red1}{RGB}{255,26,26}
\definecolor{red2}{RGB}{255,52,52}
\definecolor{red3}{RGB}{200,90,90}

\begin{table*}[ht]
  \centering \small
  \begin{tabular}{p{0.32\textwidth}p{0.3\textwidth}p{0.28\textwidth}}
    \toprule
    Question & \addsent attack that does not work & Attack that works \\
    \midrule
    {\color[rgb]{0.504,0.496,0.496}Who} {\color[rgb]{0.508,0.496,0.496}was} {\color[rgb]{0.461,0.461,0.578}Count} {\color[rgb]{0.500,0.500,0.500}of} {\color[rgb]{0.996,0.250,0.250}Melfi} & Jeff Dean was the mayor of Bracco. & Jeff Dean was the mayor of Melfi. \\
    \midrule
    {\color[rgb]{0.496,0.500,0.500}What} {\color[rgb]{0.527,0.484,0.484}country} {\color[rgb]{0.488,0.488,0.520}was} {\color[rgb]{0.996,0.250,0.250}Abhisit} {\color[rgb]{0.738,0.379,0.379}Vejjajiva} {\color[rgb]{0.594,0.453,0.453}prime} {\color[rgb]{0.605,0.445,0.445}minister} {\color[rgb]{0.496,0.500,0.500}of} {\color[rgb]{0.496,0.500,0.500},} {\color[rgb]{0.496,0.500,0.500}despite} {\color[rgb]{0.500,0.500,0.496}having} {\color[rgb]{0.500,0.500,0.500}been} {\color[rgb]{0.430,0.430,0.637}born} {\color[rgb]{0.492,0.492,0.516}in} {\color[rgb]{0.488,0.488,0.520}Newcastle} {\color[rgb]{0.500,0.500,0.496}?} & Samak Samak was prime minister of the country of Chicago, despite having been born in Leeds. & Abhisit Vejjajiva was chief minister of the country of Chicago, despite having been born in Leeds. \\
    \midrule
    {\color[rgb]{0.512,0.492,0.492}Where} {\color[rgb]{0.559,0.469,0.469}according} {\color[rgb]{0.496,0.496,0.508}to} {\color[rgb]{0.484,0.484,0.527}gross} {\color[rgb]{0.629,0.434,0.434}state} {\color[rgb]{0.531,0.484,0.484}product} {\color[rgb]{0.559,0.469,0.469}does} {\color[rgb]{0.473,0.473,0.551}Victoria} {\color[rgb]{0.996,0.250,0.250}rank} {\color[rgb]{0.547,0.477,0.477}in} {\color[rgb]{0.660,0.418,0.418}Australia} {\color[rgb]{0.508,0.496,0.496}?} & According to net state product, Adelaide ranks 7 in New Zealand & According to net state product, Adelaide ranked 7 in Australia. (as a prefix) \\
    \midrule
    {\color[rgb]{0.676,0.410,0.410}When} {\color[rgb]{0.754,0.371,0.371}did} {\color[rgb]{0.492,0.492,0.512}the} {\color[rgb]{0.461,0.461,0.574}Methodist} {\color[rgb]{0.926,0.285,0.285}Protestant} {\color[rgb]{0.480,0.480,0.535}Church} {\color[rgb]{0.625,0.438,0.438}split} {\color[rgb]{0.996,0.250,0.250}from} {\color[rgb]{0.496,0.496,0.504}the} {\color[rgb]{0.488,0.488,0.523}Methodist} {\color[rgb]{0.473,0.473,0.555}Episcopal} {\color[rgb]{0.609,0.445,0.445}Church} {\color[rgb]{0.496,0.496,0.508}?} & The Presbyterian Catholics split from the Presbyterian Anglican in 1805. & The Methodist Protestant Church split from the Presbyterian Anglican in 1805. (as a prefix) \\
    \midrule
    {\color[rgb]{0.520,0.488,0.488}What} {\color[rgb]{0.996,0.250,0.250}period} {\color[rgb]{0.504,0.496,0.496}was} {\color[rgb]{0.547,0.477,0.477}2.5} {\color[rgb]{0.477,0.477,0.547}million} {\color[rgb]{0.473,0.473,0.551}years} {\color[rgb]{0.410,0.410,0.676}ago} {\color[rgb]{0.504,0.496,0.496}?} & The period of Plasticean era was 2.5 billion years ago. & The period of Plasticean era was 1.5 billion years ago. (as a prefix) \\
    \bottomrule
  \end{tabular}
  \caption{\small \addsent attacks that failed to fool the model. With
  modifications to preserve nouns with high attributions, these are successful
  in fooling the model. Question words that receive high attribution are
  colored red (intensity indicates magnitude).}
  \label{appendix:tbl:modified_squad_attacks}
  \vspace{-3mm}
\end{table*}

\subsection{Task, model, and data}
The reading comprehension task involves identifying a span from a context paragraph as an
answer to a question. The SQuAD dataset~\cite{RajpurkarZLL16} for machine reading comprehension
contains 107.7K query-answer pairs, with 87.5K for training, 10.1K for
validation, and another 10.1K for testing. Deep learning methods are quite successful on this
problem, with the state-of-the-art F1 score at 84.6 achieved by~\newcite{YuDLLZC2018}; we analyze their model.

\subsection{Analyzing adversarial examples}\label{sec:squad:attacks}
Recall the adversarial attacks proposed by~\newcite{JL17} for reading
comprehension systems. Their attack \addsent appends sentences to the paragraph that
resemble an answer to the question without changing the ground truth. See the second column of \cref{appendix:tbl:modified_squad_attacks} for a few examples.

We investigate the effectiveness of their attacks using attributions. 
We analyze $100$ examples generated by the \addsent method in~\newcite{JL17}, and find that an adversarial sentence is successful in fooling the model in
two cases:

First, a contentful word in the question gets low/zero attribution and the
  adversarially added sentence modifies that word. E.g. in the question,
  ``Who did Kubiak take the place of after Super Bowl XXIV?'', the word
  ``Super'' gets low attribution. Adding ``After \emph{Champ} Bowl XXV,
    Crowton took the place of Jeff Dean'' changes the prediction for the model.
Second, a contentful word in the question that is not present in the context.
  For e.g. in the question ``Where hotel did the Panthers stay at?'',
    ``hotel'', is not present in the context. Adding ``The Vikings
    stayed at Chicago hotel.'' changes the prediction for the model.

%\begin{enumerate}[leftmargin=*,noitemsep]
%\item A contentful word in the question gets low/zero attribution and the
%  adversarially added sentence modifies that word. E.g. in the question,
%  \emph{Who did Kubiak take the place of after Super Bowl XXIV?}, the word
%  \emph{Super} gets low attribution. Adding \emph{After \emph{Champ} Bowl XXV,
%    Crowton took the place of Jeff Dean} changes the prediction for the model.
%\item A contentful word in the question that is not present in the context.
%  E.g. for the question \emph{Where hotel did the Panthers stay at?},
%    \emph{hotel} is not present in the context. Adding \emph{The Vikings
%    stayed at Chicago hotel.} changes the prediction for the model.
%\end{enumerate}

On the flip side, an adversarial sentence is unsuccessful when a
contentful word in the question having high attribution is not present in
the added sentence. E.g. for ``Where according to gross state
product does Victoria rank in Australia?'', ``Australia'' receives high
attribution. Adding ``According to net state product,
Adelaide ranks 7 in New Zealand.'' does not fool the model. However, retaining
``Australia'' in the adversarial sentence does change the model's prediction. 

\subsection{Predicting the effectiveness of attacks}
Next we correlate attributions with efficacy of the \addsent attacks.
We analyzed 1000 (question, attack phrase) instances\footnote{data sourced from \url{https://worksheets.codalab.org/worksheets/0xc86d3ebe69a3427d91f9aaa63f7d1e7d/} } where~\newcite{YuDLLZC2018} model has the correct baseline
prediction. Of the 1000 cases, 508 are able to fool the model,
while 492 are not. We split the examples into two groups. The first group has
examples where a noun or adjective in the question has high attribution, but
is missing from the adversarial sentence and the rest are in the second group.
Our attribution analysis suggests that we should find more failed examples in the
first group. That is indeed the case. The first group has $63\%$ failed
examples, while the second has only $40\%$.

Recall that the attack sentences were constructed by (a) generating a sentence that answers the question, (b)
replacing all the adjectives and nouns with antonyms, and named entities by the nearest word in GloVe word vector
space~\cite{PenningtonSM14} and (c) crowdsourcing to check that the new sentence is grammatically correct.
This suggests a use of attributions to improve the effectiveness of the attacks, namely ensuring that question words 
that the model thinks are important are left untouched in step (b) (we note that other changes in should be carried out).
%We would have to ensure that there is enough of a change in step (b) to ensure that the ground truth does not change.
In \cref{appendix:tbl:modified_squad_attacks}, we show a few examples where an original attack did not fool the model, but preserving a noun with
high attribution did.

\section{Conclusion}

We analyzed three question answering models using an
attribution technique. Attributions helped us identify
weaknesses of these models more effectively than conventional methods (based on validation sets).  We believe that a
workflow that uses attributions can aid the developer in iterating on
model quality more effectively.

While the attacks in this paper may seem
  unrealistic, they do expose real weaknesses that affect the
  usage of a QA product. Under-reliance on important question terms is
  not safe. We also believe that other QA models may share these weaknesses.
  Our attribution-based methods can be directly used to gauge
  the extent of such problems. Additionally, our perturbation attacks (\cref{sec:np:attacks,sec:vqa:attacks}) serve as empirical validation of attributions.

\section*{Reproducibility}
Code to generate attributions and reproduce our results is freely available at \url{https://github.com/pramodkaushik/acl18_results}.

\section*{Acknowledgments}
We thank the anonymous reviewers and Kevin Gimpel for feedback on our work, and David Dohan for helping with the reading
  comprehension network. We are grateful to Ji\v{r}\'{i} \v{S}im\v{s}a for helpful comments on drafts of this paper.

%The acknowledgments should go immediately before the references.  Do not number the acknowledgments section ({\em i.e.}, use \verb|\section*| instead of \verb|\section|). Do not include this section when submitting your paper for review.

% include your own bib file like this:
%\bibliographystyle{acl}
%\bibliography{acl2018}
\clearpage
\newpage 
\bibliography{paper}
\bibliographystyle{acl_natbib}

\appendix
\clearpage 
\newpage 
\newpage 
\section{Supplementary Material}
\label{appendix:grammar}

\begin{figure}[!htbp]
    \centering
    \emph{show,  tell, did,  me,  my, our,  are,  is,
        were, this, on, would, and,  for, should,  be, 
        do, I, have, had, the,  there,  look, give, 
        has,  was, we, get, does, a,  an, 's,  that,  
        by,  based, in,  of, bring,  with, to, from, 
        whole, being, been,  want, wanted, as, can, 
        see, doing, got, sorted, draw, listed, chart, only}
    \caption{\small Neural Programmer~\cite{NLAMA17}: List of stop words used in stop word deletion attacks (\cref{sec:np:attacks}). NP's accuracy falls from $\npaccuracy\%$ to $28.5\%$ on deleting stop words from questions in the validation set.}
    \label{fig:fluffwords}
\end{figure} 

\begin{table*}[!hbp]
  \centering \small
\begin{tabular}{ll}
    \toprule
    Operator &                                                          Triggers \\
    \midrule
    \opwordmatch &     [tm\_token, many, how, number, or, total, after, before, only] \\
    \opprev &              [before, {\color{red} many}, {\color{red} than}, previous, above, {\color{red}how}, {\color{red}at}, {\color{red}most}] \\
    \opfirst &         [tm\_token, first, before, {\color{red} after}, {\color{red} who}, previous, {\color{red} or}, peak] \\
    \opresetselect &           [many, total, how, number, last, least, the, first, of] \\
    \opcount &  [many, how, number, total, {\color{red} of}, difference, between, long, times] \\
    \opnext &              [after, {\color{red} not}, {\color{red} many}, next, {\color{red} same}, tm\_token, {\color{red} how}, below] \\
    \oplast &                [last, {\color{red} or}, after, tm\_token, next, {\color{red} the}, {\color{red} chart}, {\color{red} not}] \\
    \opgroupbymax &                                            [most, cm\_token, {\color{red} same}] \\
    \opmin &                                                 [least, {\color{red} the}, {\color{red} not}] \\
    \opmax &                                                   [most, largest] \\
    \opgeq &                  [at, more, least, {\color{red} had}, over, number, than, many] \\
    \opprint &                                                        [tm\_token] \\
    \bottomrule
\end{tabular}

\caption{\small Neural Programmer~\cite{NLAMA17}: Operator triggers. Notice that there are several irrelevant triggers (highlighted in red). For instance, ``many'' is irrelevant to 	``\opprev''. See Section~\ref{sec:np:attacks} for attacks exploiting this weakness.}
\label{appendix:tbl:operatortriggers}
\end{table*}

\begin{table*}[!hbp]
    \centering \small
\begin{tabular}{lrl}
    \toprule
    Operator sequence &  \#tables &                                                                                     Triggers \\
    \midrule
    \opresetselect, \opresetselect, \opmax, \opprint &      109 &                     [\unktoken, date, position, points, name, competition, notes, no, year, venue] \\
    \opresetselect, \opprev, \opmax, \opprint &       68 &                             [\unktoken, rank, total, bronze, gold, silver, nation, name, date, no] \\
    \opresetselect, \opresetselect, \opfirst, \opprint &       29 &              [name, \unktoken, notes, year, nationality, rank, location, date, comments, hometown] \\
    \opresetselect, \opgroupbymax, \opfirst, \opprint &       25 &                         [notes, date, title, \unktoken, role, genre, year, score, opponent, event] \\
    \opresetselect, \opresetselect, \opmin, \opprint &       17 &                            [year, height, \unktoken, name, position, floors, notes, jan, jun, may] \\
    \opresetselect, \opgroupbymax, \opmax, \opprint &       14 &             [opponent, date, result, location, rank, site, attendance, notes, city, listing] \\
    \opresetselect, \opnext, \opfirst, \opprint &       10 &                             [\unktoken, name, year, edition, birth, death, men, time, women, type] \\
    \opresetselect, \opresetselect, \oplast, \opprint &        9 &              [date, \unktoken, distance, location, name, year, winner, japanese, duration, member] \\
    \opresetselect, \opprev, \opfirst, \opprint &        7 &  [name, notes, intersecting, kilometers, location, athlete, nationality, rank, time, design] \\
    \opresetselect, \opnext, \opmax, \opprint &        7 &                                                                             [\unktoken, ethnicity] \\
    \opresetselect, \opgroupbymax, \oplast, \opprint &        6 &             [place, season, \unktoken, date, division, tier, builder, cylinders, notes, withdrawn] \\
    \opresetselect, \opprev, \oplast, \opprint &        5 &                  [report, date, average, chassis, driver, race, builder, name, notes, works] \\
    \opresetselect, \opprev, \opmin, \opprint &        4 &          [division, level, movements, position, season, current, gauge, notes, wheel, works] \\
    \opresetselect, \opwordmatch, \oplast, \opprint &        3 &                           [car, finish, laps, led, rank, retired, start, year, \unktoken, carries] \\
    \opresetselect, \opresetselect, \opfirst, \opcount &        2 &                                                           [\unktoken, network, owner, programming] \\
    \opresetselect, \opresetselect, \opresetselect, \opprint &        1 &                                      [candidates, district, first, incumbent, party, result] \\
    \opresetselect, \opwordmatch, \opwordmatch, \opprint &        1 &                                                [lifetime, name, nationality, notable, notes] \\
    \opresetselect, \opresetselect, \oplast, \opcount &        1 &                                                         [\unktoken, english, japanese, type, year] \\
    \opresetselect, \opnext, \oplast, \opprint &        1 &                                                                               [\unktoken, comment] \\
    \opresetselect, \opgroupbymax, \opwordmatch, \opprint &        1 &                                                    [\unktoken, length, performer, producer, title] \\
    \bottomrule
\end{tabular}
    \caption{\small Neural Programmer~\cite{NLAMA17}: Column attributions in occurring in table-specific default programs, classified by operator sequence.  These attributions are indication that the network is predisposed towards picking certain operators solely based on the table. Here is an insight: the second row of this table indicates that the network is inclined to choosing ``\opresetselect, \opprev'' on tables about medal tallies. It may have learned this bias as some medal tables in training data may have ``\emph{total}'' rows which may confound answer computation if not excluded by applying ``\opresetselect, \opprev''. However, not all medal tables have ``\emph{total}'' rows, and hence ``\opresetselect, \opprev'' must not be applied universally. As the operators are triggered by column names and not by table entries, the network cannot distinguish between tables with and without ``\emph{total}'' row and may erroneously exclude the last rows.}
    \label{appendix:tbl:default_programs}

\end{table*}

\end{document}